\documentclass{article}
\usepackage{ijcai23}

\usepackage{times}
\usepackage{soul}
\usepackage{url}
\usepackage[hidelinks]{hyperref}
\usepackage[utf8]{inputenc}
\usepackage[small]{caption}
\usepackage{graphicx}
\usepackage{amsmath}
\usepackage{amsthm}
\usepackage{booktabs}
\usepackage{algorithm}
\usepackage{algorithmicx}
\usepackage[switch]{lineno}
\usepackage{algpseudocode}
\usepackage{array}
\usepackage{amssymb}
\usepackage{arydshln}
\usepackage{amsthm,amsmath,amssymb}
\usepackage{mathrsfs}

\title{Contrastive Label Enhancement}

\author{
    Yifei Wang
    \and
    Yiyang Zhou\and
    Jihua Zhu\thanks{Corresponding author}\and
    Xinyuan Liu\and
    Wenbiao Yan\And
    Zhiqiang Tian
    \affiliations
    School of Software Engineering, Xi’an Jiaotong University, Xi’an, China\\
    \emails
    \{wangyf.ailab,zhouyiyangailab\}@gmail.com,
    zhujh@xjtu.edu.cn,
    \{xinyuan.liu,wenbiao777\}@stu.xjtu.edu.cn,
    zhiqiangtian@xjtu.edu.cn
}
\begin{document}

\maketitle

\begin{abstract}
% Label distribution learning (LDL) is a new machine learning paradigm for solving label ambiguity. Since it is difficult to directly obtain label distributions, many studies are focusing on how to recover label distributions from logical labels, dubbed label enhancement (LE). This paper proposes a novel method called Contrast Label Enhancement (ConLE) to consider features and logical labels as sample descriptions in different dimensions. They are fully integrated to generate high-level features, which are further used to obtain label distributions for each sample. This is different from previous LE methods, which directly build a mapping relationship between features and label distributions under the supervision of logical labels. Specifically, ConLE learns feature representations to transform both features and logical labels of one sample into the unified projection space by the contrastive learning strategy, which makes features and logical labels of the same sample closer, while features and labels of different samples are farther away in projection space. Subsequently, the obtained high-level features are mapped into label distributions by the feature mapping network, where its training is guided by a well-designed strategy considering the consistency of label attributes. Extensive experiments on benchmark datasets demonstrate the effectiveness and superiority of our method.

Label distribution learning (LDL) is a new machine learning paradigm for solving label ambiguity. Since it is difficult to directly obtain label distributions, many studies are focusing on how to recover label distributions from logical labels, dubbed label enhancement (LE). Existing LE methods estimate label distributions by simply building a mapping relationship between features and label distributions under the supervision of logical labels. They typically overlook the fact that both features and logical labels are descriptions of the instance from different views. Therefore, we propose a novel method called Contrastive Label Enhancement (ConLE) which integrates features and logical labels into the unified projection space to generate high-level features by contrastive learning strategy. In this approach, features and logical labels belonging to the same sample are pulled closer, while those of different samples are projected farther away from each other in the projection space. Subsequently, we leverage the obtained high-level features to gain label distributions through a well-designed training strategy that considers the consistency of label attributes. Extensive experiments on LDL benchmark datasets demonstrate the effectiveness and superiority of our method.

% Label distribution learning (LDL) is a novel machine learning paradigm that uses to solve label ambiguity. Since label distributions are difficult to obtain, more research is focusing on how to recover label distributions from logical labels, which is called label enhancement (LE). Current LE methods mainly pay attention to building a mapping relationship between feature and label distribution and using logical labels to supervise the recovery of label distribution. However, in this paper, we propose a new method called Contrast Label Enhancement (CLE). The outstanding contribution of this method is that, unlike the existing LE methods, it considers features and logical labels as two similar descriptions of the same sample, and skillfully incorporates them to generate high-level features of samples. Specifically, we learn a representation learning model through the contrastive learning strategy, which makes the features and logical labels of the same sample closer in projection space, while the features and labels of different samples are farther away. Moreover, we design a training strategy considering the consistency of label attributes to guide the learning of feature mapping networks. Extensive experiments on 13 benchmark datasets demonstrate the effectiveness and superiority of our methods. 
\end{abstract}

\section{Introduction}
\begin{figure}[htbp]
    \centering
    \includegraphics[scale=0.4]{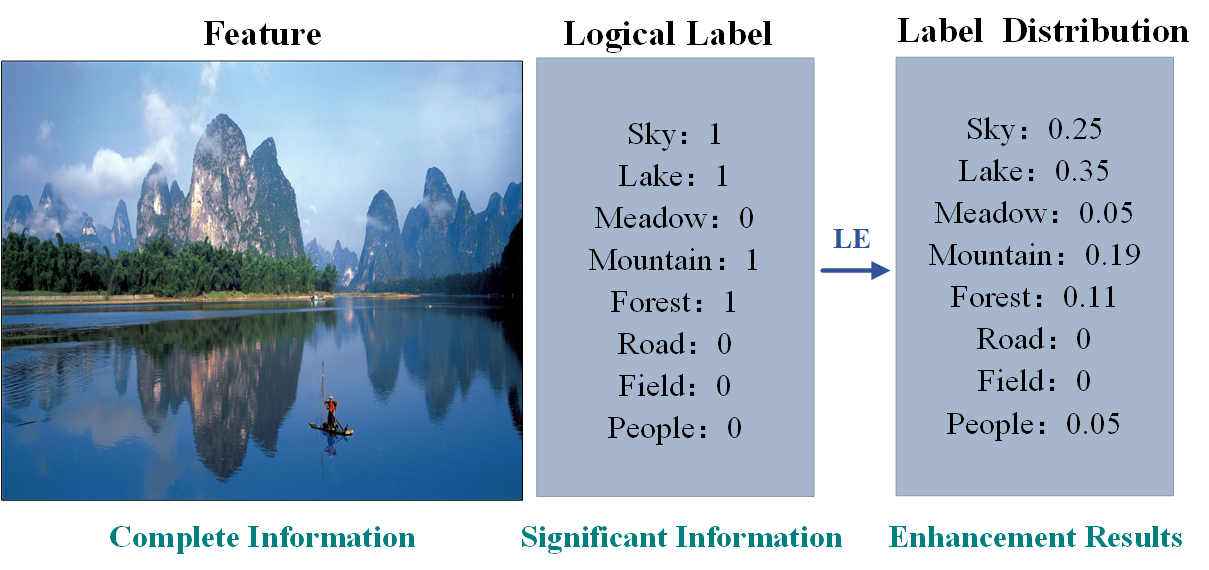}
    \caption{ An example of label enhancement. Features contain the full information of samples with many redundancies, while logical labels possess significant information but are not comprehensive. The generation of label distributions makes full use of the important knowledge in logical labels and supplements the sample details according to the features. } \label{fig:1}
    \end{figure} 
In recent years, Label Distribution Learning (LDL) \cite{geng2016label} has drawn much attention in machine learning, with its effectiveness demonstrated in various applications \cite{geng2013facial,zhang2015crowd,qi2022label}. Unlike single-label learning (SLL) and multi-label learning (MLL) \cite{gibaja2014multi,moyano2019evolutionary,ijcai2022p524}, LDL can provide information on how much each label describes a sample, which helps to deal with the problem of label ambiguity \cite{geng2016label}. However, Obtaining label distributions is more challenging than logical labels, as it requires many annotators to manually indicate the degree to which each label describes an instance and accurately quantifying this degree remains difficult. Thus, \cite{xu2019label} proposed Label Enhancement (LE), leveraging the topological information in the feature space and the correlation among the labels to recover label distributions from logical labels.

More specifically, LE can be seen as a preprocessing of LDL \cite{zheng2021generalized}, which takes the logically labeled datasets as inputs and outputs label distributions. As shown in Figure\,\ref{fig:1}, this image reflects the complete information of the sample including some details. Meanwhile, its corresponding logical labels only highlight the most salient features, such as the sky, lake, mountain, and forest. Features contain comprehensive information about samples with many redundancies, while logical labels hold arresting information but are not all-sided. Therefore, it is reasonable to assume that features and logical labels can be regarded as two descriptions of instances from different views, possessing complete and salient information of samples. The purpose of LE tasks can be simplified as enhancing the significant knowledge in logical labels by utilizing detailed features. Subsequently, each label is allocated a descriptive degree according to its importance.

Most existing LE methods concentrate on establishing the mapping relationship between features and label distributions under the guidance of logical labels. Although these previous works have achieved good performance for LE problem, they neglect that features and labels are descriptions of two different dimensions related to the same samples. Furthermore, logical labels can only indicate the conspicuous information of each sample without obtaining the label description ranking. The label distributions may appear to be quite different even if the logical labels present the same results.

To address these issues, we propose the ConLE method which fuses features and logic labels to generate the high-level features of samples by contrastive learning strategy. More specifically, we elaborately train a representation learning model, which forces the features and logical labels of the same instance to be close in projection space, while those of different instances are farther away. By concatenating the representations of features and logical labels in projection space, we get high-level features including knowledge of logic labels and features. Accordingly, label distributions can be recovered from high-level features by the feature mapping network. Since it is expected that the properties of labels in the recovered label distributions should be consistent with those in the logical labels, we design a training strategy with label-level consistency to guide the learning of the feature mapping network.

Our contributions can be delivered as follows:

\begin{itemize}

% \item By analyzing the problem of label enhancement, we regard features and logical labels as descriptions from different views, and take full advantage of their intrinsic relevance to propose the Contrastive Label Enhancement (ConLE) method, which transforms features and logical labels into the unified projection space to generate high-level features for label enhancement.

\item Based on our analysis of label enhancement, we recognize that features and logical labels offer distinct perspectives on instances, with features providing comprehensive information and logical labels highlighting salient information. In order to leverage the intrinsic relevance between these two views, we propose the Contrastive Label Enhancement (ConLE) method, which unifies features and logical labels in a projection space to generate high-level features for label enhancement.

\item Since all possible labels should have similar properties in logical labels and label distributions, we design a training strategy to keep the consistency of label properties for the generation of label distributions. This strategy not only maintains the attributes of relevant and irrelevant labels but also minimizes the distance between logical labels and label distributions.

\item Extensive experiments are conducted on 13 benchmark datasets, experimental results validate the effectiveness and superiority of our ConLE compared with several state-of-the-art LE methods.

\end{itemize}

\begin{figure*}[ht]
    \centering
    \includegraphics[scale=0.62]{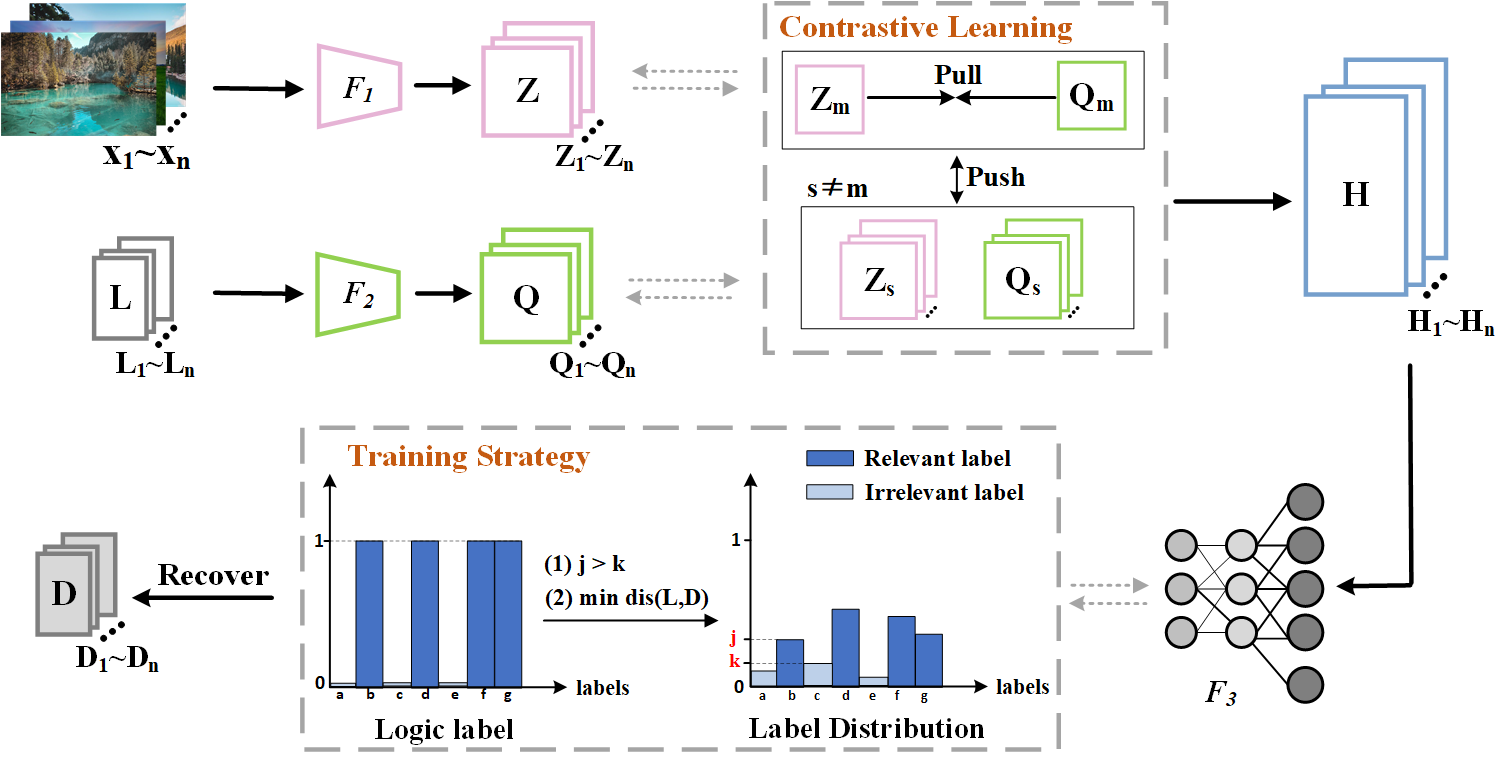}
    \caption{
    Framework of the proposed ConLE. ConLE approaches the LE problem by regarding features ($X$) and logical labels ($L$) as sample descriptions from two views. It uses two mapping networks ($F_1$ and $F_2$) to project $X$ and $L$ into a unified projection space, which results in two representations $Z$ and $Q$. These representations are then concatenated into high-level features ($H$). To obtain good high-level features, ConLE utilizes a contrastive learning strategy that brings two representations of the same sample closer together while pushing representations of different samples farther apart from each other. Additionally, ConLE employs a reliable training strategy to generate label distributions $D$ from high-level features $H$ by the feature mapping network $F_3$. This strategy minimizes the distance between logical labels and label distributions, ensuring that the restored label distributions are close to the existing logical labels. Meanwhile, it also demands the description degree of relevant labels marked as 1 in the logical labels is larger than that of the irrelevant labels marked as 0. In this way, ConLE can guarantee the consistency of label attributes in logical labels and label distributions.} \label{fig:2}
    \end{figure*} 
    
    % The framework of our ConLE model for label enhancement. By investigating the fundamental thought of the LE algorithm, ConLE views features ($X$) and logical labels ($L$) as sample descriptions of two distinct dimensions. In the first step, we utilize two mapping networks ($F1$ and $F2$) to project $X$ and $L$ into the unified space, resulting in two representations $Z1$ and $Z2$.  We adopt contrastive learning strategy for fetching these two descriptions of identical sample closer and alienating them of different samples, taking the final representations as high-level features ($H$). In the second step, we predominantly focus on the process of generating label distributions $D$ from high-level features $H$. For such a need, we present a trustworthy training strategy to guarantee the consistency of labels attributes in logical labels and label distributions. This strategy has two components. The first part is used to minimize the distance between logical labels and label distributions, and the second part pays attention to ensure that the label description degree of relevant labels marked as 1 in the logical labels is larger than that of the irrelevant labels marked as 0.

\section{Related Work}
In this section, we mainly introduce the related work of this paper from two research directions: label enhancement and contrastive learning.

\paragraph{{Label Enhancement.}} Label enhancement is proposed to recover label distributions from logical labels and provide data preparation for LDL. For example, the Graph Laplacian LE (GLLE) method proposed by \cite{8868206} makes the learned label distributions close to logical labels while accounting for learning label correlations, making similar samples have similar label distributions. The method LESC proposed by \cite{tang2020label} uses low-rank representations to excavate the underlying information contained in the feature space. \cite{xu2022variational} proposed LEVI to infer label distributions from logical labels via variational inference. The method RLLE formulates label enhancement as a dynamic decision process and uses prior knowledge to define the target for LE \cite{gao2021label}. The kernel-based label enhancement (KM) algorithm maps each instance to a high-dimensional space and uses a kernel function to calculate the distance between samples and the center of the group, in order to obtain the label description. \cite{jiang2006fuzzy}. The LE algorithm based on label propagation (LP) recovers label distributions from logical labels by using the iterative label propagation technique \cite{li2015leveraging}. Sequential label enhancement (Seq\_LE) formulates the LE task as a sequential decision procedure, which is more consistent with the process of annotating the label distributions in human brains \cite{gao2022sequential}. However, these works neglect the essential connection between features and logical labels. In this paper, we regard features and logical labels as sample descriptions from different views, where we can create faithful high-level features for label enhancement by integrating them into the unified projection space.
~

% text \noindent\textbf{Label Enhancement.}\quad 

\paragraph{{Contrastive Learning.}}The basic idea of contrastive learning, an excellent representation learning method, is to map the original data to a feature space. Within this space, the objective is to maximize the similarities among positive pairs while minimizing those among negative pairs. \cite{grill2020bootstrap,li2020prototypical}. Currently, contrastive learning has achieved good results in many machine learning domains \cite{li2021contrastive,dai2017contrastive}. Here we primarily introduce several contrastive learning methods applied to multi-label learning. \cite{wang2022contrastive} designed a multi-label contrastive learning objective in the multi-label text classification task, which improves the retrieval process of their KNN-based method. \cite{zhang2022use} present a hierarchical multi-label representation learning framework that can leverage all available labels and preserve the hierarchical relationship between classes. \cite{9815553} propose two novel models to learn discriminative and modality-invariant representations for cross-modal retrieval. \cite{bai2022gaussian} propose a novel contrastive learning boosted multi-label prediction model based on a Gaussian mixture variational autoencoder (C-GMVAE), which learns a multimodal prior space and employs a contrastive loss. For ConLE, the descriptions of one identical sample are regarded as positive pairs and those of different samples are negative pairs. We pull positive pairs close and negative pairs farther away in projection space by contrastive learning to obtain good highlevel features, which is really beneficial for the LE process.

\section{The ConLE Approach}
In this paper, we use the following notations. The set of instances is denoted by $X = \{x_1,x_2,...,x_n\} \in \mathbb{R}^{dim_1 \times n}$, where $dim_1$ is the dimensionality of each instance and $n$ is the number of instances. $Y = \{y_1, y_2,..., y_c\}$ denotes the complete set of labels, where $c$ is the number of classes. For an instance $x_i$, its logical label is represented by $L_i = (l_{x_i}^{y_1}, l_{x_i}^{y_2}, \dots, l_{x_i}^{y_c})\textsuperscript{T}$, where $l_{x_i}^{y_j}$ can only take values of 0 or 1.
The label distribution for $x_i$ is denoted by $D_i = (d_{x_i}^{y_1}, d_{x_i}^{y_2}, \dots, d_{x_i}^{y_c})\textsuperscript{T}$, where $d_{x_i}^{y_j}$ depicts the degree to which $x_i$ belongs to label $y_j$. It is worth noting that the sum of all label description degrees for $x_i$ is equal to 1. The purpose of LE tasks is to recover the label distribution $D_i$ of $x_i$ from the logical label $L_i$ and transform the logically labeled dataset $S = \{(x_i, L_i) | 1 \leq i \leq n\}$ into the LDL training set $E = \{(x_i, D_i) | 1 \leq i \leq n\}$.
The proposed Contrastive Label Enhancement (ConLE) in this paper contains two important components: the generation of high-level features by contrastive learning and the training strategy with label-level consistency for LE. Overall, the loss function of ConLE can be formulated as follows:
\begin{equation}
L_{ConLE} = l_{con} + l_{att}.
\label{for:1}
\end{equation}
where $l_{con}$ denotes the contrastive loss for high-level features, $l_{att}$ indicates the loss of training strategy with label-level consistency. The framework of ConLE and the detailed procedure of these two parts is shown in Figure\,\ref{fig:2}. 

\subsection{The Generation of High-Level Features by Contrastive Learning}
The first section provides a detailed analysis of the essence of LE tasks. We regard features and logic labels as two descriptions of samples. Features contain complete information, while logic labels capture prominent details. Label distributions show the description degree of each label. We can't simply focus on the salient information in logical labels, but make good use of salient information and supplement the detailed information according to the original features. To effectively excavate the knowledge of features and logical labels, we adopt the contrastive learning of sample-level consistency. 

To reduce the information loss induced by contrastive loss, we do not directly conduct contrastive learning on the feature matrix \cite{li2021contrastive}. Instead, we project the features ($X$) and logical labels ($L$) of all samples into a unified projection space via two mapping networks ($F_1(\cdot;\theta)$,$F_2(\cdot;\phi)$), and then get the representations $Z$ and $Q$. Specifically, the representations of features and logic labels in the projection space can be obtained by the following formula:
\begin{equation}
Z_m = F_1(x_m;\theta),
\label{for:2}
\end{equation}
\begin{equation}
Q_m = F_2(L_m;\phi),
\label{for:3}
\end{equation}
where $x_m$ and $L_m$ represent the features and logical labels of the m-th sample, $Z_m$ and $Q_m$ denote their embedded representations in the $dim_2$-dimensional space. $\theta$ and $\phi$ refer to the corresponding network parameters.

Contrastive learning aims to maximize the similarities of positive pairs while minimizing those of negative ones. In this paper, we construct positive and negative pairs at the instance level with Z and Q where $\{Z_m,Q_m\}$ is positive pair and leave other $(n-1)$ pairs to be negative. The cosine similarity is utilized to measure the closeness degree between pairs:
\begin{equation}
h(Z_m,Q_m)=\frac{(Z_m)(Q_m)\textsuperscript{T}}{||Z_m||\,||Q_m||}.
\end{equation}

To optimize pairwise similarities without losing their generality, the form of instance-level contrastive loss between $Z_m$ and $Q_m$ is defined as:
\begin{equation}
l_m = l_{Z_m}+l_{Q_m},
\end{equation}
where $l_{Z_m}$ denotes the contrastive loss for $Z_m$ and $l_{Q_m}$ indicates loss of $Q_m$. Specifically, the item $l_{Z_m}$ is defined as:
\begin{equation}
l_{Z_m} = -log\frac{e^{(h(Z_m,Q_m)/\tau_I)}}{ {\textstyle \sum_{s=1,s\neq{m}}^{n}[e^{(h(Z_m,Z_s)/\tau_I)}+e^{(h(Z_m,Q_s)/\tau_I)}]} }, 
\label{for:6}
\end{equation}
and the item $l_{Q_m}$ is formulated as:
\begin{equation}
l_{Q_m} = -log\frac{e^{(h(Q_m,Z_m)/\tau_I)}}{ {\textstyle \sum_{s=1,s\neq{m}}^{n}[e^{(h(Q_m,Q_s)/\tau_I)}+e^{(h(Q_m,Z_s)/\tau_I)}]} },
\label{for:7}
\end{equation}
where $\tau_I$ is the instance-level temperature parameter to control the softness. Further, the instance-level contrastive loss is computed across all samples as:
\begin{equation}
l_{con} = \frac{1}{n} {\textstyle \sum_{m=1}^{n}}l_{m}. 
\label{for:8}
\end{equation}

The expressions $Z$ and $Q$ updated by contrastive learning strategy will be concatenated as high-level features $H$, which are taken as inputs of the feature mapping network to learn the label distributions:
\begin{equation}
H=concat(Z,Q).
\label{for:9}
\end{equation}

% \subsection{The training strategy for label-level consistency}
\subsection{The Training Strategy With Label-Level Consistency for LE}
Based on the obtained high-level features, we introduce a feature mapping network $F_3$ to generate label distributions. In other words, we have the following formula:
\begin{equation}
D_m = F_3(H_m;\varphi), 
\label{for:10}
\end{equation}
where $D_m$ is the recovered label distribution of the m-th sample and $H_m$ is the high-level feature, and $\varphi$ denote the parameter of feature mapping network $F_3$ .

In ConLE, we consider the consistency of label attributes in logical labels and label distributions. Firstly, because of recovered label distributions should be close to existing logical labels, we expect to minimize the distance between logical labels and the recovered label distributions, which is normalized by the softmax normalization form. This criterion can be defined as:
\begin{equation}
l_{dis} = \sum_{m=1}^{n} {||F_3(H_m;\varphi) -L_m||}^2,
\end{equation}
where $D_m$ and $L_m$ represents the recovered label distribution and logic label of the m-th sample.
Moreover, logical labels divide all possible labels into relevant labels marked 1 and irrelevant labels marked 0 for each sample. We hope to ensure that the attributes of relevant and irrelevant labels are consistent in label distributions and logical labels. This idea is considered in many multi-label learning methods \cite{kanehira2016multi,yan2016multi}. Under their inspiration, we apply a threshold strategy to ensure that the description degree of relevant labels should be greater than that of irrelevant labels in the recovered label distributions. This strategy can be written as follows:
\begin{equation}
\begin{aligned}
& d^{y^+}_{x_m} - d^{y^-}_{x_m} > 0 \\
& \text{s.t.}\quad y^+ \in P_m, y^- \in N_m 
\end{aligned}
\end{equation}
where $P_m$ is used to indicate the set of relevant labels in $x_m$, $N_m$ represents the set of irrelevant labels in $x_m$, $d^{y^+}_{x_m}$ and $d^{y^-}_{x_m}$ are the prediction results of LE process. 

In this way, we can get the loss function of threshold strategy:
% \begin{equation}
% l_{thr} = \frac{1}{n} \sum_{m=1}^{n} \sum_{L^-_m} \sum_{L^+_m} [\max (d^{L^-_m}_{x_m}-d^{L^+_m}_{x_m}+\alpha,0)]
% \end{equation}
\begin{equation}
l_{thr} = \frac{1}{n} \sum_{m=1}^{n} \sum_{\scriptstyle y^+ \in P_m} \sum_{\scriptstyle y^- \in N_m} [\max (d^{y^-}_{x_m}-d^{y^+}_{x_m}+\epsilon,0)],
\end{equation}
where $\epsilon$ is a hyperparameter that determines the threshold. The formula can be simplified to:
\begin{equation}
l_{thr} = \frac{1}{n} {\textstyle \sum_{m=1}^{n}} [\max (\max d^{y^-}_{x_m}-\min d^{y^+}_{x_m}+\epsilon,0)],
\end{equation}

Finally, the loss function of training strategy for label-level consistency can be formulated as follows:
\begin{equation}
l_{att} = \lambda _1l_{dis} + \lambda _2l_{thr},
\label{for:15}
\end{equation}
where $\lambda _1$ and $\lambda _1$ are two trade-off parameters.

This designed training strategy can guarantee that label attributes are the same in the logical labels and label distributions, thus obtaining a better feature mapping network to recover label distributions. The full optimization process of ConLE is summarized in Algorithm \ref{alg:1}.
% LE methods usually take the fitting degree between the recovered label distributions and the real label distributions as the evaluation standard.

% \begin{algorithm}[t]
%     \caption{The optimization of ConLE}
%     \label{alg:1}
%     \hspace*{0.02in} {\bf Input:} %算法的输入， \hspace*{0.02in}用来控制位置，同时利用 \\ 进行换行
%     Training instances $X=[X_1,X_2,...,X_n]$; Logical labels $L=\{L_1,L_2,...,L_n\}$; Temperature parameter $\tau_I$\\
%     \hspace*{0.02in} {\bf Output:} %算法的结果输出
%     label distributions $D=\{D_1,D_2,...,D_n\}$
%     \begin{algorithmic}[1]
%         \State Initialize ${\theta}$ by minimizing Eq.\,(\ref{for:6});
%         \State Initialize ${\phi}$ by minimizing Eq.\,(\ref{for:7});
%         \State Initialize $H$ by Eq.\,(\ref{for:9});
%         \While{not converged}
%             \State  Optimize ${\theta}$, ${\phi}$ by Eq.\,(\ref{for:7}) and Eq.\,(\ref{for:8});
%             \State Obtain the high-level features $H$ by Eq.\,(\ref{for:9});
%             \State Update $\varphi$, $D$ through Eq.\,(\ref{for:15});
%         \EndWhile
%         \State \Return $D$
%     \end{algorithmic}
% \end{algorithm}

\begin{algorithm}[t]
    \caption{The optimization of ConLE}
    \label{alg:1}
    \hspace*{0.05in} {\bf Input:} %算法的输入， \hspace*{0.02in}用来控制位置，同时利用 \\ 进行换行
    Training instances $X=\{x_1,x_2,...,x_n\}$; Logical labels $L=\{L_1,L_2,...,L_n\}$; Temperature parameter $\tau_I$\\
    \hspace*{0.02in} {\bf Output:} %算法的结果输出
    label distributions $D=\{D_1,D_2,...,D_n\}$
    \begin{algorithmic}[1]
        \State Random Initialize ${\theta}$, ${\phi}$ and $\varphi$;
        % \State Initialize $\{Z_m, Q_m\}^n_{m=1}$ by Eq.\,(\ref{for:2}) and Eq.\,(\ref{for:3});
        \While{not converged}
            \State Obtain $\{Z_m, Q_m\}^n_{m=1}$ by Eq.\,(\ref{for:2}) and Eq.\,(\ref{for:3});
            \State Obtain the high-level features $H$ by Eq.\,(\ref{for:9});
            \State Obtain label distributions $D$ by Eq.\,(\ref{for:10});
            \State Optimize ${\theta}$, ${\phi}$, $\varphi$ through Eq.\,(\ref{for:1});
        \EndWhile
        \State \Return $D$
    \end{algorithmic}
\end{algorithm}

\section{Experiments}
% In this section, experiments are conducted on 13 real-world datasets with 7 advanced LE methods to verify the effectiveness of our method. We select six criteria that are most commonly used for performance evaluation. All methods are implemented by PyTorch.
\subsection{Datasets}
We conduct comprehensive experiments on 13 real-world datasets to verify the effectiveness of our method. To be specific, SJAFFE dataset \cite{lyons1998coding} and SBU-3DFE dataset \cite{yin20063d} are obtained from the two facial expression databases, JAFFE and BU-3DFE. Each image in datasets is rated for six different emotions (i.e., happiness, sadness, surprise, fear, anger, and disgust) using 5-level scale. The Natural Scene dataset is collected from 2000 natural scene images. Dataset Movie is about the user rating for 7755 movies. Yeast datasets are derived from biological experiments on gene expression levels of budding yeast at different time points \cite{eisen1998cluster}. The basic statistics of these datasets are shown in Table \ref{ta:1}.

\begin{table}[t]
\fontsize{9pt}{11pt}
\renewcommand\arraystretch{1}
\centering
    \begin{tabular}{ccccc}
    \toprule %添加表格头部粗线
    No. & Dataset & Examples & Features & Labels \\
    \midrule %第二道横线 
    1 & SJAFFE & 213 & 243 & 6 \\ 
    2 & SBU-3DFE & 2500 & 243 & 6 \\ 
    3 & Natural-Scene & 2000 & 294 & 9 \\ 
    4 & Movie & 7755 & 1869 & 5 \\ 
    5 & Yeast-alpha & 2465 & 24 & 18 \\ 
    6 & Yeast-cdc & 2465 & 24 & 15 \\ 
    7 & Yeast-elu & 2465 & 24 & 14 \\ 
    8 & Yeast-diau & 2465 & 24 & 7 \\ 
    9 & Yeast-dtt & 2465 & 24 & 4 \\
    10 & Yeast-heat & 2465 & 24 & 6 \\ 
    11 & Yeast-cold & 2465 & 24 & 4 \\ 
    12 & Yeast-spo & 2465 & 24 & 6 \\ 
    13 & Yeast-spo5 & 2465 & 24 & 3 \\ 
    \bottomrule %第三道横线
    \end{tabular}
\caption{Statistics of the 13 datasets.}
\label{ta:1}
\end{table}

\subsection{Evaluation Measures}
The performance of the LE algorithm is usually calculated by distance or similarity between the recovered label distributions and the real label distributions. According to \cite{geng2016label}, we select six measures to evaluate the recovery performance, i.e., Kullback-Leibler divergence (K-L)↓, Chebyshev distance (Cheb)↓, Clark distance (Clark)↓, Canberra metric (Canber)↓, Cosine coefficient (Cosine)↑ and Intersection similarity (Intersec)↑. The first four are distance measures and the last two are similarity measures. The formulae for these six measures are summarized in Table \ref{ta:2}.

\begin{table}[t]
\small
\renewcommand\arraystretch{2.3}
\centering
    \begin{tabular}{cc}
    \toprule %添加表格头部粗线
    Measure & Formula \\
    \midrule %第二道横线 
    \rule{0pt}{10pt}
    Kullback-Leibler↓ & $Dis_1(D,\hat{D})=\sum_{j=1}^cd_jln\frac{d_j}{\hat{d_j}}$ \\
    \rule{0pt}{10pt}
    Chebyshev↓ & $Dis_2(D,\hat{D})=max_{j}|d_j-\hat{d_j}|$ \\
    \rule{0pt}{10pt}
    Clark↓ & $Dis_3(D,\hat{D})=\sqrt{\sum_{j=1}^c\frac{{(d_j-\hat{d_j})}^2}{{(d_j+\hat{d_j})}^2}}$ \\ 
    \rule{0pt}{10pt}
    Canberra↓ & $Dis_4(D,\hat{D})=\sum_{j=1}^c\frac{|d_j-\hat{d_j}|^2}{d_j+\hat{d_j}}$ \\
    \rule{0pt}{10pt}
    Cosine↑ & $Sim_1(D,\hat{D})=\frac{\sum_{j=1}^cd_j\hat{d_j}}{\sqrt{\sum_{j=1}^cd_j^2}\sqrt{\sum_{j=1}^c\hat{d_j}^2}}$ \\
    \rule{0pt}{10pt}
    Intersection↑ & $Sim_2(D,\hat{D})=\sum_{j=1}^cmin(d_j,\hat{d_j})$ \\
    \bottomrule %第三道横线
    \end{tabular}
\caption{Introduction to evalution measures.}
\label{ta:2}
\end{table}

% We compare CLE against seven LE methods, including FCM, KM, LP, GLLE, LEVI, LESC and $L^2$-LE. For all experiments with CLE, the temperature parameter $\tau_I$ is fixed to 0.5, the trade-off parameters $\lambda _1$ and $\lambda _1$ are set to 1. In order to ensure the fairness of experiments, all other comparison algorithms are based on the code shared by the authors and all of them adopt the default settings as recommended in original papers.

% To investigate the recovery performance, quantitative results for the above algorithm were presented in six metrics (as shown in Table III)
\begin{table*}[ht]
\centering  % 显示位置为中间
\scriptsize 
% \fontsize{9pt}{1pt}\selectfont
\resizebox{\linewidth}{!}{
\begin{tabular}{r|ccccccc|ccccccc}
\hline
Metrics      & \multicolumn{7}{c|}{Kullback-Leibler ↓}                                                                          & \multicolumn{7}{c}{Chebyshev ↓}                                                                                       \\ 
Methods  & \multicolumn{1}{l}{FCM} & \multicolumn{1}{l}{KM} & \multicolumn{1}{l}{LP} & \multicolumn{1}{l}{GLLE} & \multicolumn{1}{l}{LEVI-MLP} & \multicolumn{1}{l}{LESC} & \multicolumn{1}{l|}{ConLE} & \multicolumn{1}{l}{FCM} & \multicolumn{1}{l}{KM} & \multicolumn{1}{l}{LP} & \multicolumn{1}{l}{GLLE} & \multicolumn{1}{l}{LEVI-MLP} & \multicolumn{1}{l}{LESC} & \multicolumn{1}{l}{ConLE} \\ \hline
SJAFFE           & 0.107                  & 0.558               & 0.077                      & 0.050                         &  0.031                  & 0.029                                      & \textbf{0.028}                       &  0.132                      & 0.214                      &  0.107                       & 0.087         & 0.073                & \textbf{0.069}                        & \textbf{0.069}                             \\
SBU-3DFE           & 0.199                       & 0.583                       &  0.108                      & 0.069                         &  0.042                   & 0.064                                   & \textbf{0.039}                        & 0.230                     & 0.234                      &  0.161                      & 0.122               &  0.092                       & 0.122                                     & \textbf{0.082}                           \\
Natural-Scene                         & 3.131                     &  3.009                          & 1.680                   & 2.663                  &  0.928                 & 1.166                      & \textbf{0.757}                     & 0.368                       &  0.306                      &  \textbf{0.275}              &0.335                      & 0.324                    &  0.341                   & 0.314   \\
Movie            &  0.381                      & 0.452                       &  0.177                    &  0.123                           & 0.081                   &   0.120                  & \textbf{0.060}                 & 0.230                       &  0.234                      &  0.161                     &  0.122                       & 0.109               &  0.121                       & \textbf{0.097}                            \\
Yeast-alpha        & 0.100                       &  0.630                       & 0.121                      & 0.013           & 0.006             &  0.008             & \textbf{0.005}                  & 0.044                       &  0.063                      & 0.040                       & 0.020                    & \textbf{0.013}                       & 0.015                     & \textbf{0.013}                           \\
Yeast-cdc                   & 0.091                     &  0.530                  &0.013                    & 0.014                   &  0.006                 &  0.010                        &  \textbf{0.004}                    &  0.051                       &  0.076                       & 0.042              & 0.022                       & 0.015                     & 0.019                       & \textbf{0.014}                           \\
Yeast-elu                              &  0.059                      &  0.609                          &  0.012                  &  0.013                   & 0.007                   & 0.009                        & \textbf{0.004}                      & 0.052                      & 0.078                     & 0.044               &  0.023                       &  0.017                       &  0.019                     & \textbf{0.013}                           \\
Yeast-diau                & 0.159                     &0.538                 & 0.127                   & 0.027                   & 0.011                   & 0.017                       & \textbf{0.009}                      & 0.124                       &  0.152                       &  0.099               &  0.053                      &  0.033                       & 0.042                      & \textbf{0.031}                            \\
Yeast-dtt                 & 0.065                       & 0.617                 & 0.103                   &  0.013                   & 0.011                   &  0.010                       & \textbf{0.009}                      & 0.097                      &  0.257                       &  0.128               &  0.052                      &  0.051                     &  \textbf{0.043}                       & 0.047                          \\
Yeast-heat                    & 0.147                      &  0.586                          &  0.089                   &  0.017                  &  0.008                   &  0.015                      & \textbf{0.007}                       & 0.169                      &  0.175                       &  0.086               &  0.049                      & 0.033                      &  0.046                       & \textbf{0.031}                           \\
Yeast-cold                       &0.113                       &  0.586                           & 0.103                   & 0.019                   & 0.011                   & 0.015                       & \textbf{0.009}                       & 0.141                      &  0.252                       &0.137               &  0.066                      & 0.051                         & 0.056                       & \textbf{0.044}                           \\
Yeast-spo                               & 0.110                      &  0.562                           &  0.084                  & 0.029                   &0.014                    &   0.028                      & \textbf{0.013}                       &  0.130                       &  0.175                      & 0.090               &  0.062                       &  0.045                       &  0.060                      & \textbf{0.042}                            \\
Yeast-spo5                           &0.123                      &  0.334                           & 0.042                   & 0.034                   & 0.015                    & 0.031                       & \textbf{0.013}                     & 0.162                       & 0.277                       & 0.114              &0.099                      &  0.067                      & 0.092                     & \textbf{0.060}                           \\
 \cdashline{1-15}[1pt/1pt]
Avg.Rank  &5.92        & 6.92           &4.92       & 4.23                        & 2.15                            &2.84                         &\textbf{1.00}                         & 6.00                        &6.61    &4.53   &4.15    &2.30    & 3.07   & \textbf{1.23}                                          \\ \hline

Metrics            & \multicolumn{7}{c|}{Clark ↓}                                                                          & \multicolumn{7}{c}{Canberra ↓}                                                                                       \\ 
Datasets  & \multicolumn{1}{l}{FCM} & \multicolumn{1}{l}{KM} & \multicolumn{1}{l}{LP} & \multicolumn{1}{l}{GLLE} & \multicolumn{1}{l}{LEVI-MLP} & \multicolumn{1}{l}{LESC}  & \multicolumn{1}{l|}{ConLE} & \multicolumn{1}{l}{FCM} & \multicolumn{1}{l}{KM} & \multicolumn{1}{l}{LP} & \multicolumn{1}{l}{GLLE} & \multicolumn{1}{l}{LEVI-MLP} & \multicolumn{1}{l}{LESC} & \multicolumn{1}{l}{ConLE} \\ \hline
SJAFFE           & 0.522                  & 1.874               & 0.451                      & 0.377                         & 0.285                    & 0.276                                 & \textbf{0.269}                       &   1.081                      &  4.010                      &   1.064                       &  0.781         & 0.587                & 0.561                                    & \textbf{0.545}                             \\
SBU-3DFE           & 0.482                      & 1.907                      & 0.580                       & 0.391                           & 0.304                   &  0.378                                      & \textbf{0.297}                         & 1.020                      & 4.121                      &  1.245                       & 0.828             &    
    \textbf{0.635}                      &  0.799                                      & 0.670                            \\
Natural-Scene                         & 2.486                     &  2.448                          &2.482                 & 2.460                  & 2.454                 & 2.464                       & \textbf{2.450}                     &  6.974                     & 6.795                     &   6.790           &6.851                      & 6.801                     &  6.878                 & \textbf{6.708}   \\
Movie            &  0.859                    &  1.766                      & 0.913                      &  0.569                      &  0.548                    & 0.564                   & \textbf{0.463}                   & 1.664                        &  3.444                      &  1.720                     & 1.045                                      &  0.968                      &  1.034                     & \textbf{0.837}                            \\
Yeast-alpha        & 0.821                       &  3.153                     &  1.185                       &  0.337                          &  0.219                   & 0.253                   &  \textbf{0.214}                                  & 2.883                     & 11.809                       &  4.544              &  1.134                      & 0.732                       & 0.846                     & \textbf{0.696}                            \\
Yeast-cdc            &  0.739                      & 2.885                      & 1.014                                 & 0.306                   &  0.209                   & 0.251                       & \textbf{0.178}                       &  2.415                       & 9.875                       &  3.644              &  0.959                      & 0.642                      &  0.765                     & \textbf{0.505}                           \\
Yeast-elu              &0.579                       & 2.768                       & 0.973                      & 0.295                          &  0.222                   & 0.241                   &\textbf{0.165}                    & 1.689                       &  9.110                       &  3.381                      & 0.902                      &  0.674              & 0.727                      & \textbf{0.480}                                       \\
Yeast-diau                                & 0.838                      &  1.886                           &  0.788                   & 0.296                   & 0.191                   & 0.224                      & \textbf{0.175}                       & 1.895                      & 4.261                       &  1.748              &  0.671            & 0.421                    & 0.480                        & \textbf{0.365}                            \\
Yeast-dtt                           & 0.329                       & 1.477                         &  0.499                  &  0.143                   & 0.140                   & 0.119                       & \textbf{0.114}                      & 0.501                       & 2.594                       & 0.941             & 0.248                      & 0.247                       & 0.206                       & \textbf{0.199}                            \\
Yeast-heat                   & 0.580                       &  1.802                          &  0.568                   & 0.213                   &  0.147                   &  0.199                       & \textbf{0.136}                       & 1.157                       &  3.849                       & 1.293              & 0.430                       &  0.295                       & 0.401                      & \textbf{0.268}                           \\
Yeast-cold                      & 0.433                       & 1.472                           & 0.503                   & 0.176                   &  0.140                   &   0.152                      &\textbf{0.119}                       & 0.734                      & 2.566                       & 0.924               &  0.305                       &  0.243                       &0.263                       & \textbf{0.203}                            \\
Yeast-spo                            & 0.520                       & 1.811                          &  0.558                   &  0.266                   &  0.187                   &  0.258                       & \textbf{0.177}                       & 0.998                       &  3.854                       &  1.231               & 0.548                       & 0.372                      &0.533                       & \textbf{0.353}                          \\
Yeast-spo5         & 0.395                    &  1.059                      &   0.274               & 0.197                   &  0.136                   &   0.185            & \textbf{0.127}                       & 0.563                     &  1.382                    & 0.401          & 0.305                    &   0.208             & 0.284                      & \textbf{0.192}                     \\ \cdashline{1-15}[1pt/1pt]
Avg.Rank  & 5.15       & 6.84                        & 5.53                        & 4.07                        & 2.30                            & 3.07                        & \textbf{1.00}                        & 5.30                         & 6.69  &5.53  & 4.07
& 2.15 & 3.07 & \textbf{1.07}                                               \\ \hline

Metrics            & \multicolumn{7}{c|}{Cosine ↑}                                                                          & \multicolumn{7}{c}{Intersection ↑}                                                                                       \\ 
Datasets  & \multicolumn{1}{l}{FCM} & \multicolumn{1}{l}{KM} & \multicolumn{1}{l}{LP} & \multicolumn{1}{l}{GLLE} & \multicolumn{1}{l}{LEVI-MLP} & \multicolumn{1}{l}{LESC}  & \multicolumn{1}{l|}{ConLE} & \multicolumn{1}{l}{FCM} & \multicolumn{1}{l}{KM} & \multicolumn{1}{l}{LP} & \multicolumn{1}{l}{GLLE} & \multicolumn{1}{l}{LEVI-MLP} & \multicolumn{1}{l}{LESC} & \multicolumn{1}{l}{ConLE} \\ \hline
SJAFFE           &  0.906                  & 0.827               & 0.941                      & 0.958                                          & \textbf{0.973}                    & 0.970                  & 0.972                       &  0.821                      & 0.593                      &  0.837                       & 0.872         & 0.899                & 0.905                                     & \textbf{0.907}                     
\\
SBU-3DFE           & 0.912                     & 0.812                      & 0.922                      & 0.927                        & 0.957                   &  0.932                                & \textbf{0.963}    &  0.827                     &  0.579                      & 0.810                   & 0.850                      &  0.882             & 0.855                                       & \textbf{0.886}                                            
\\
Natural-Scene                         & 0.593                     &  0.748                          & \textbf{0.860}               & 0.778                  & 0.712                  & 0.760                      & 0.804                     &  0.312                  & 0.416                     &   0.451           &0.522                    & 0.441                   &  0.510                 & \textbf{0.537}   \\
Movie            & 0.773                      & 0.880                      &  0.929                       & 0.936                                     &  0.955                   & 0.937                       & \textbf{0.964}                       & 0.677                      &  0.649                      & 0.778              & 0.831                      &  0.850                       & 0.833                      & \textbf{0.871}                          \\
Yeast-alpha                  & 0.922                       & 0.751               & 0.911                    &   0.987                  &  \textbf{0.995}                  & 0.992                       & \textbf{0.995}                      & 0.844                       &  0.532                     & 0.774              &  0.938                       &  0.960               &  0.953                        &\textbf{0.961}                           \\
Yeast-cdc                 & 0.929                       & 0.754                           &  0.916                   & 0.987                   & 0.994                   &   0.991                      & \textbf{0.995}                     & 0.847                       & 0.533                     &  0.779              & 0.937                      &  0.958                       & 0.950                       & \textbf{0.966}                           \\
Yeast-elu                            &  0.950                       &0.758                          & 0.918                   &  0.987                   &  0.993                   &  0.991                       & \textbf{0.996}                     & 0.883                       &0.539                      &0.782              & 0.936                       & 0.952                       & 0.949                       & \textbf{0.966}                          \\
Yeast-diau                        & 0.882                       &  0.799                           &  0.915                   & 0.975                   &  0.990                   &  0.985                      & \textbf{0.991}                       & 0.760                       & 0.588                      & 0.788               &  0.906                       &  0.942                     & 0.933                      & \textbf{0.949}                           \\
Yeast-dtt                            &0.959                     & 0.759                          &  0.921                   &   0.988                  &0.990                    & 0.991                       & \textbf{0.992}                      &0.894                       &  0.541                      & 0.786               & 0.939                       &  0.939                      &  0.949                    & \textbf{0.950}                           \\
Yeast-heat                       &0.883                       &0.779                           &0.932                    &0.984                    &  0.992                  & 0.986                        & \textbf{0.993}                      & 0.807                      &  0.559                       &0.805               &  0.929                       &  0.952                      & 0.934                       & \textbf{0.956}                           \\
Yeast-cold                      & 0.922                       & 0.779                           & 0.925                   & 0.982                   &  0.990                   &  0.986                       & \textbf{0.991}                       & 0.833                       & 0.559                       &  0.794               & 0.924                      &  0.940                      &  0.935                      & \textbf{0.950}                            \\
Yeast-spo                                 & 0.909                      & 0.800                           & 0.939                  & 0.974                   &  0.988                &  0.975                       &\textbf{0.989}                       & 0.836                      & 0.575                       & 0.819               & 0.909                       & 0.940                      &  0.912                      & \textbf{0.942}                          \\
Yeast-spo5          & 0.922                   &  0.882                        &  0.969              & 0.971             & 0.987        &   0.974                 & \textbf{0.988}                   & 0.838                    & 0.724                    & 0.886            &  0.901                     & 0.933                  & 0.908                    & \textbf{0.939}                      \\ \cdashline{1-15}[1pt/1pt]
Avg.Rank  & 5.84       & 6.30                 & 4.76                        & 4.00                        &2.00                             & 2.76                        & \textbf{1.15}                        & 5.46  & 6.92 &5.46 &3.92 &2.38 &2.92 & \textbf{1.00}                                              \\ 
\bottomrule
\end{tabular}}
\caption{Recovery results evaluated by six measures.} 
\label{ta:3}
\end{table*}

\subsection{Comparison Methods}
We compare ConLE with six advanced LE methods, including FCM \cite{gayar2006study}, KM \cite{jiang2006fuzzy}, LP \cite{li2015leveraging}, GLLE \cite{8868206}, LEVI-MLP \cite{xu2022variational} and LESC \cite{tang2020label}. The following are the datails of comparison algorithms used in our experiments:

\textbf{1) FCM:} This method makes use of membership degree to determine which cluster each instance belongs to according to fuzzy C-means clustering.

\textbf{2) KM:} It is a kernel-based algorithm that uses the fuzzy SVM to get the radius and center, obtaining the membership degree as the final label distribution. 

\textbf{3) LP:} This approach applies label propagation (LP) in semi-supervised learning to label enhancement, employing graph models to construct a label propagation matrix and generate label distributions.

\textbf{4) GLLE:} The algorithm recovers label distributions in the feature space guided by the topological information.

\textbf{5) LEVI-MLP:} It regards label distributions as potential vectors and infers them from the logical labels in the training datasets by using variational inference.

\textbf{6) LESC:} This method utilizes the low-rank representation to capture the global relationship of samples and predict implicit label correlation to achieve label enhancement.

\begin{table*}[ht]
\centering  % 显示位置为中间
\scriptsize
% \fontsize{8pt}{9pt}\selectfont
% \resizebox{\linewidth}{!}{
\begin{tabular}{r|ccc|ccc|ccc|ccc}
\hline
Metrics  & \multicolumn{3}{c|}{Kullback-Leibler ↓}                                                                          & \multicolumn{3}{c|}{Clark ↓}   & \multicolumn{3}{c|}{Canberra ↓}  & \multicolumn{3}{c}{Intersection ↑}                                                                                       \\ 
Methods  & \multicolumn{1}{l}{ConLE$_h$} & \multicolumn{1}{l}{ConLE$_l$} & \multicolumn{1}{l|}{ConLE} & \multicolumn{1}{l}{ConLE$_h$} & \multicolumn{1}{l}{ConLE$_l$} & \multicolumn{1}{l|}{ConLE} & \multicolumn{1}{l}{ConLE$_h$} & \multicolumn{1}{l}{ConLE$_l$} & \multicolumn{1}{l|}{ConLE} & \multicolumn{1}{l}{ConLE$_h$} & \multicolumn{1}{l}{ConLE$_l$} &\multicolumn{1}{l}{ConLE} \\ \hline
SJAFFE           & 0.399                  & 0.044               & \textbf{0.028}                      & 0.320                         &  0.305         & \textbf{0.269}       & 0.651  & 0.713 & \textbf{0.545}     & 0.888       & 0.892  & \textbf{0.907}\\
SBU-3DFE           & 0.051                  & 0.060               & \textbf{0.039}                      & 0.365                         &  0.405         & \textbf{0.297}       & 0.767  & 0.850 & \textbf{0.670}     & 0.867       & 0.842  & \textbf{0.886}\\
Natural-Scene           & 0.795                  & 0.773               & \textbf{0.757}                      & 2.463                         &  2.443         & \textbf{2.450}       & 6.802  & 6.695 & \textbf{6.708}     & 0.497       & 0.503  & \textbf{0.537}\\
Movie           & 0.073                  & 0.068               & \textbf{0.060}                      & 0.517                         &  0.491         & \textbf{0.463}       & 0.923  & 0.877 & \textbf{0.837}     & 0.858       & 0.866  & \textbf{0.871}\\
Yeast-alpha           & 0.007                  & 0.010               & \textbf{0.005}                      & 0.244                         &  0.342         & \textbf{0.214}       & 0.728  & 0.799 & \textbf{0.696}     & 0.920       & 0.891  & \textbf{0.961}\\
Yeast-cdc           & 0.006                  & 0.006               & \textbf{0.004}                      & 0.210                         &  0.231         & \textbf{0.178}       & 0.618  & 0.609 & \textbf{0.505}     & 0.959       & 0.960  & \textbf{0.966}\\
Yeast-elu           & 0.006                  & 0.007               & \textbf{0.004}                      & 0.199                         &  0.204         & \textbf{0.165}       & 0.582  & 0.599 & \textbf{0.480}     & 0.959       & 0.955  & \textbf{0.966}\\
Yeast-diau           & 0.018                  & 0.014               & \textbf{0.009}                      & 0.248                         &  0.198         & \textbf{0.175}       & 0.509  & 0.405 & \textbf{0.365}     & 0.930       & 0.937  & \textbf{0.949}\\
Yeast-dtt           & 0.013                  & 0.015               & \textbf{0.009}                      & 0.156                         &  0.201         & \textbf{0.114}       & 0.298  & 0.349 & \textbf{0.199}     & 0.942       & 0.930  & \textbf{0.950}\\
Yeast-heat           & 0.016                  & 0.012               & \textbf{0.007}                      & 0.302                         &  0.267         & \textbf{0.136}       & 0.412  & 0.370 & \textbf{0.268}     & 0.929       & 0.941  & \textbf{0.956}\\
Yeast-cold           & 0.012                  & 0.011               & \textbf{0.009}                      & 0.190                         &  0.162         & \textbf{0.119}       & 0.331  & 0.286 & \textbf{0.203}     & 0.939       & 0.931  & \textbf{0.950}\\
Yeast-spo           & 0.019                  & 0.016               & \textbf{0.013}                      & 0.285                         &  0.246         & \textbf{0.177}       & 0.443  & 0.406 & \textbf{0.353}     & 0.914       & 0.927  & \textbf{0.942}\\
Yeast-spo5           & 0.014                  & 0.015               & \textbf{0.013}                      & 0.157                         &  0.172         & \textbf{0.127}       & 0.248  & 0.230 & \textbf{0.192}     & 0.923       & 0.929  & \textbf{0.939}\\
\bottomrule
\end{tabular}
\caption{Recovery results of ConLE$_h$, ConLE$_l$ and ConLE on 13 real-world datasets.} 
\label{ta:4}
\end{table*}

\begin{figure*}[ht]
    \centering
    \includegraphics[scale=0.425]{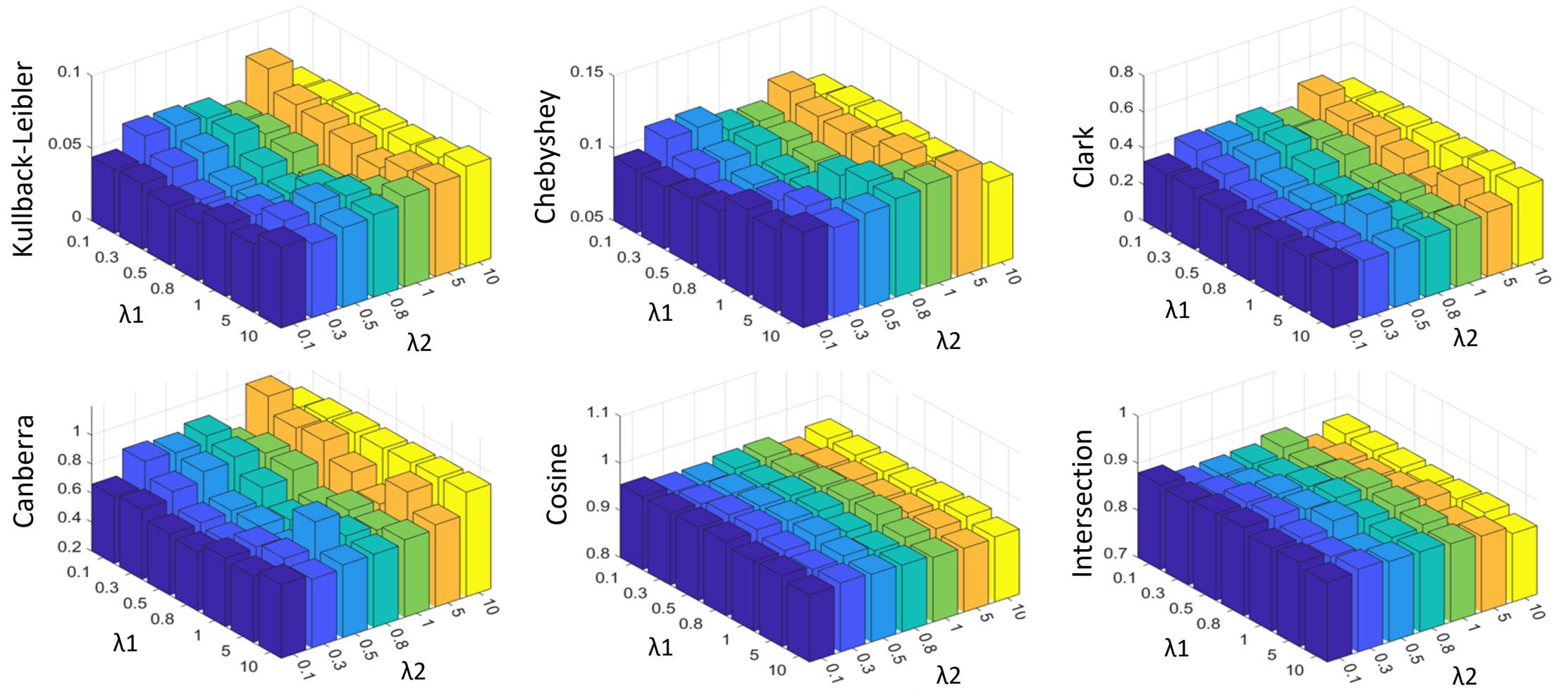}
    \caption{Influence of parameters $\lambda _1$ and $\lambda _2$ on dataset SBU-3DFE.} \label{fig:3}
    \end{figure*} 

\begin{figure}[ht]
    \centering
    \includegraphics[scale=0.34]{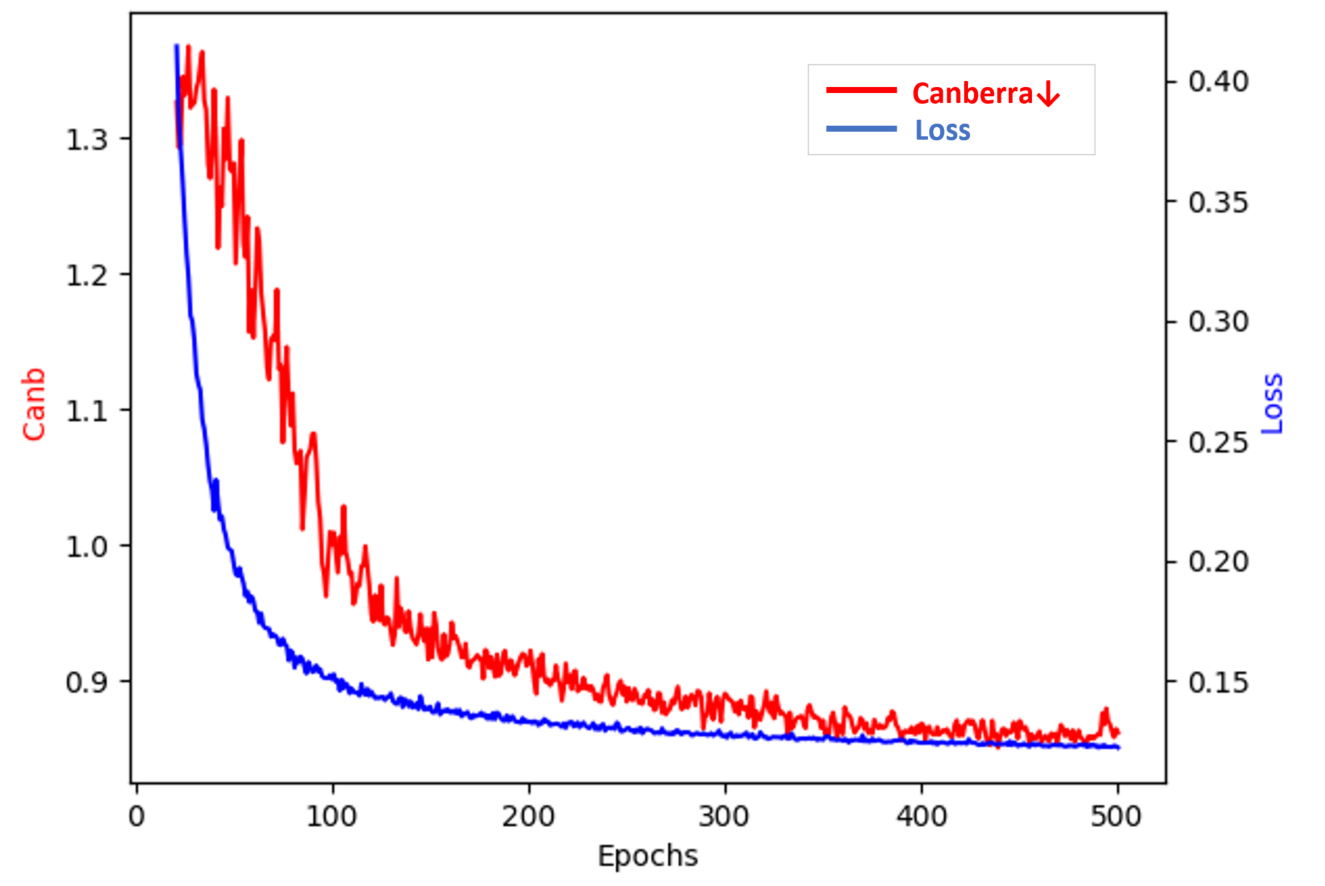}
    \caption{Convergence curve on dataset Movie.} \label{fig:4}
    \end{figure}

\subsection{Experimental Results}
\paragraph{Implementation Details.}In ConLE, we adopt the SGD optimizer \cite{ruder2016overview} for optimization and utilize the LeakyReLU activation function \cite{maas2013rectifier} to implement the networks. The code of this method is implemented by PyTorch \cite{paszke2019pytorch} on one NVIDIA Geforce GTX 2080ti GPU with 11GB memory. All experiments for our selected comparison algorithms follow the optimal settings mentioned in their papers and we run the programs using the code provided by their relevant authors. All algorithms are evaluated by ten times ten-fold cross-validation for fairness. When comparing with other algorithms, the hyperparameters of ConLE are set as follows: $\lambda_1$ is set to 0.5, $\lambda_2$ is set to 1 and the temperature parameter $\tau_I$ is 0.5.
\paragraph{Recovery Performance.}
The detailed comparison results are presented in Table \ref{ta:3}, with the best performance on each dataset highlighted in bold. For each evaluation metric, ↓ shows the smaller the better while ↑ shows the larger the better. The average rankings of each algorithm across all the datasets are shown in the last row of each table. 

The experimental results clearly indicate that our ConLE method exhibits superior recovery performance compared to the other six advanced LE algorithms. Specifically, ConLE can achieve the ranking of 1.00, 1.23, 1.00, 1.07, 1.15 and 1.00 respectively for the six evaluation metrics. ConLE obtains excellent performance both on large-scale datasets such as movie and small-scale datasets such as SJAFFE. ConLE can attain significant improvements both in comparison with algorithm adaption and specialized algorithms by exploring the description consistency of features and logical labels in the same sample. We integrate features and logical labels into the unified projection space to generate high-level features and keep the consistency of label attributes in the process of label enhancement.
\paragraph{Ablation Studies.} 
Our ConLE method consists of two main components: generating high-level features by contrastive learning and a training strategy with label-level consistency for LE. Ablation studies are conducted to verify the effectiveness of the two modules in our method.

Therefore, we first remove the part of ConLE that generates high-level features and get a comparison algorithm ConLE$_h$, whose loss function can be written as:
\begin{equation}
L_{ConLE_h} = \lambda _1l_{dis} + \lambda _2l_{thr},
\end{equation}
In ConLE$_h$, we only explore the consistency information of label attributes without considering the description consistency of features and labels.

Secondly, we need to remove the strategy that ensures the consistency of label attributes. To ensure the normal training process, we still keep the strategy of minimizing the distance between label distributions and logical labels. The loss function of the comparison function ConLE$_l$:
\begin{equation}
L_{ConLE_l} = \lambda _1l_{dis} + l_{con}.
\end{equation}

Table \ref{ta:4} provides the recovery results of ConLE$_h$, ConLE$_l$ and ConLE. Due to the limitation of space, only the representative results measured on Kullback-Leibler, Clark, Canberra and Intersection are shown in the table. From the experimental results, we can observe that ConLE is superior to ConLE$_h$ and ConLE$_l$ in all cases. Compared with ConLE$_h$, ConLE considers the inherent relationship between features and logical labels. It grasps the description consistency of samples and constructs high-level features for training. Compared with ConLE$_l$, ConLE considers label-level consistency of logical labels and label distributions. It makes that each relevant label in the logical labels has a greater description degree in the label distributions. Therefore, our experimental results have verified that both modules of ConLE play essential roles in achieving excellent recovery performance. The integration of these modules in the complete ConLE method has been demonstrated to be highly effective.
\paragraph{Parameters Sensitivity.} 
To investigate the sensitivity of ConLE to hyperparameters, we performed experiments on SBU-3DFE with different values of the two trade-off hyperparameters $\lambda _1$ and $\lambda _2$. In this experiment, we fix one hyperparameter and choose another hyperparameter from \{0.1, 0.3, 0.5, 0.8 ,1, 5, 10\}. As shown in  Figure \ref{fig:3}, we can observe that the ConLE method can obtain satisfactory recovery results and our model is insensitive to $\lambda _1$ and $\lambda _2$. 
\paragraph{Convergence Analysis.} 
To illustrate the convergence of ConLE, we present an experiment conducted on Movie dataset by Canberra↓ as an example, with the corresponding convergence curve depicted in Figure \ref{fig:4}. The value of the objective function decreases and the performance increases with more iterations. Finally, they tend to be stable. The properties remain the same for all datasets.

\section{Conclusion}
In this work, we propose Contrastive Label Enhancement (ConLE), a novel method to cope with the (Label Enhancement) LE problem. ConLE regards features and logic labels as descriptions from different views, and then elegantly integrates them to generate high-level features by contrastive learning. Additionally, ConLE employs a training strategy that considers the consistency of label attributes to estimate the label distributions from high-level features. Experimental results on 13 datasets demonstrate its superior performance over other state-of-the-art methods.

\section*{Acknowledgments}
This work was supported by the National Key R\&D Program of China under Grant 2020AAA0109602.

\bibliographystyle{named}
\bibliography{ijcai23}

\begin{thebibliography}{}

\bibitem[\protect\citeauthoryear{Bai \bgroup \em et al.\egroup
  }{2022}]{bai2022gaussian}
Junwen Bai, Shufeng Kong, and Carla~P Gomes.
\newblock Gaussian mixture variational autoencoder with contrastive learning
  for multi-label classification.
\newblock In {\em International Conference on Machine Learning}, pages
  1383--1398. PMLR, 2022.

\bibitem[\protect\citeauthoryear{Dai and Lin}{2017}]{dai2017contrastive}
Bo~Dai and Dahua Lin.
\newblock Contrastive learning for image captioning.
\newblock {\em Advances in Neural Information Processing Systems}, 30, 2017.

\bibitem[\protect\citeauthoryear{Eisen \bgroup \em et al.\egroup
  }{1998}]{eisen1998cluster}
Michael~B Eisen, Paul~T Spellman, Patrick~O Brown, and David Botstein.
\newblock Cluster analysis and display of genome-wide expression patterns.
\newblock {\em Proceedings of the National Academy of Sciences},
  95(25):14863--14868, 1998.

\bibitem[\protect\citeauthoryear{Gao \bgroup \em et al.\egroup
  }{2021}]{gao2021label}
Yongbiao Gao, Yu~Zhang, and Xin Geng.
\newblock Label enhancement for label distribution learning via prior
  knowledge.
\newblock In {\em Proceedings of the Twenty-Ninth International Conference on
  International Joint Conferences on Artificial Intelligence}, pages
  3223--3229, 2021.

\bibitem[\protect\citeauthoryear{Gao \bgroup \em et al.\egroup
  }{2022}]{gao2022sequential}
Yongbiao Gao, Ke~Wang, and Xin Geng.
\newblock Sequential label enhancement.
\newblock {\em IEEE Transactions on Neural Networks and Learning Systems},
  2022.

\bibitem[\protect\citeauthoryear{Gayar \bgroup \em et al.\egroup
  }{2006}]{gayar2006study}
Neamat~El Gayar, Friedhelm Schwenker, and G{\"u}nther Palm.
\newblock A study of the robustness of knn classifiers trained using soft
  labels.
\newblock In {\em IAPR Workshop on Artificial Neural Networks in Pattern
  Recognition}, pages 67--80. Springer, 2006.

\bibitem[\protect\citeauthoryear{Geng \bgroup \em et al.\egroup
  }{2013}]{geng2013facial}
Xin Geng, Chao Yin, and Zhi-Hua Zhou.
\newblock Facial age estimation by learning from label distributions.
\newblock {\em IEEE transactions on pattern analysis and machine intelligence},
  35(10):2401--2412, 2013.

\bibitem[\protect\citeauthoryear{Geng}{2016}]{geng2016label}
Xin Geng.
\newblock Label distribution learning.
\newblock {\em IEEE Transactions on Knowledge and Data Engineering},
  28(7):1734--1748, 2016.

\bibitem[\protect\citeauthoryear{Gibaja and Ventura}{2014}]{gibaja2014multi}
Eva Gibaja and Sebasti{\'a}n Ventura.
\newblock Multi-label learning: a review of the state of the art and ongoing
  research.
\newblock {\em Wiley Interdisciplinary Reviews: Data Mining and Knowledge
  Discovery}, 4(6):411--444, 2014.

\bibitem[\protect\citeauthoryear{Grill \bgroup \em et al.\egroup
  }{2020}]{grill2020bootstrap}
Jean-Bastien Grill, Florian Strub, Florent Altch{\'e}, Corentin Tallec, Pierre
  Richemond, Elena Buchatskaya, Carl Doersch, Bernardo Avila~Pires, Zhaohan
  Guo, Mohammad Gheshlaghi~Azar, et~al.
\newblock Bootstrap your own latent-a new approach to self-supervised learning.
\newblock {\em Advances in neural information processing systems},
  33:21271--21284, 2020.

\bibitem[\protect\citeauthoryear{Jiang \bgroup \em et al.\egroup
  }{2006}]{jiang2006fuzzy}
Xiufeng Jiang, Zhang Yi, and Jian~Cheng Lv.
\newblock Fuzzy svm with a new fuzzy membership function.
\newblock {\em Neural Computing \& Applications}, 15(3):268--276, 2006.

\bibitem[\protect\citeauthoryear{Kanehira and Harada}{2016}]{kanehira2016multi}
Atsushi Kanehira and Tatsuya Harada.
\newblock Multi-label ranking from positive and unlabeled data.
\newblock In {\em Proceedings of the IEEE conference on computer vision and
  pattern recognition}, pages 5138--5146, 2016.

\bibitem[\protect\citeauthoryear{Li \bgroup \em et al.\egroup
  }{2015}]{li2015leveraging}
Yu-Kun Li, Min-Ling Zhang, and Xin Geng.
\newblock Leveraging implicit relative labeling-importance information for
  effective multi-label learning.
\newblock In {\em 2015 IEEE International Conference on Data Mining}, pages
  251--260. IEEE, 2015.

\bibitem[\protect\citeauthoryear{Li \bgroup \em et al.\egroup
  }{2020}]{li2020prototypical}
Junnan Li, Pan Zhou, Caiming Xiong, and Steven~CH Hoi.
\newblock Prototypical contrastive learning of unsupervised representations.
\newblock {\em arXiv preprint arXiv:2005.04966}, 2020.

\bibitem[\protect\citeauthoryear{Li \bgroup \em et al.\egroup
  }{2021}]{li2021contrastive}
Yunfan Li, Peng Hu, Zitao Liu, Dezhong Peng, Joey~Tianyi Zhou, and Xi~Peng.
\newblock Contrastive clustering.
\newblock In {\em Proceedings of the AAAI Conference on Artificial
  Intelligence}, volume~35, pages 8547--8555, 2021.

\bibitem[\protect\citeauthoryear{Lyons \bgroup \em et al.\egroup
  }{1998}]{lyons1998coding}
Michael Lyons, Shigeru Akamatsu, Miyuki Kamachi, and Jiro Gyoba.
\newblock Coding facial expressions with gabor wavelets.
\newblock In {\em Proceedings Third IEEE international conference on automatic
  face and gesture recognition}, pages 200--205. IEEE, 1998.

\bibitem[\protect\citeauthoryear{Maas \bgroup \em et al.\egroup
  }{2013}]{maas2013rectifier}
Andrew~L Maas, Awni~Y Hannun, Andrew~Y Ng, et~al.
\newblock Rectifier nonlinearities improve neural network acoustic models.
\newblock In {\em Proc. icml}, volume~30, page~3. Atlanta, Georgia, USA, 2013.

\bibitem[\protect\citeauthoryear{Moyano \bgroup \em et al.\egroup
  }{2019}]{moyano2019evolutionary}
Jose~M Moyano, Eva~L Gibaja, Krzysztof~J Cios, and Sebasti{\'a}n Ventura.
\newblock An evolutionary approach to build ensembles of multi-label
  classifiers.
\newblock {\em Information Fusion}, 50:168--180, 2019.

\bibitem[\protect\citeauthoryear{Paszke \bgroup \em et al.\egroup
  }{2019}]{paszke2019pytorch}
Adam Paszke, Sam Gross, Francisco Massa, Adam Lerer, James Bradbury, Gregory
  Chanan, Trevor Killeen, Zeming Lin, Natalia Gimelshein, Luca Antiga, et~al.
\newblock Pytorch: An imperative style, high-performance deep learning library.
\newblock {\em Advances in neural information processing systems}, 32, 2019.

\bibitem[\protect\citeauthoryear{Qi \bgroup \em et al.\egroup
  }{2022}]{qi2022label}
Lei Qi, Jiaying Shen, Jiaqi Liu, Yinghuan Shi, and Xin Geng.
\newblock Label distribution learning for generalizable multi-source person
  re-identification.
\newblock {\em arXiv preprint arXiv:2204.05903}, 2022.

\bibitem[\protect\citeauthoryear{Qian \bgroup \em et al.\egroup
  }{2022}]{9815553}
Shengsheng Qian, Dizhan Xue, Quan Fang, and Changsheng Xu.
\newblock Integrating multi-label contrastive learning with dual adversarial
  graph neural networks for cross-modal retrieval.
\newblock {\em IEEE Transactions on Pattern Analysis and Machine Intelligence},
  pages 1--18, 2022.

\bibitem[\protect\citeauthoryear{Ruder}{2016}]{ruder2016overview}
Sebastian Ruder.
\newblock An overview of gradient descent optimization algorithms.
\newblock {\em arXiv preprint arXiv:1609.04747}, 2016.

\bibitem[\protect\citeauthoryear{Tang \bgroup \em et al.\egroup
  }{2020}]{tang2020label}
Haoyu Tang, Jihua Zhu, Qinghai Zheng, Jun Wang, Shanmin Pang, and Zhongyu Li.
\newblock Label enhancement with sample correlations via low-rank
  representation.
\newblock In {\em Proceedings of the AAAI Conference on Artificial
  Intelligence}, volume~34, pages 5932--5939, 2020.

\bibitem[\protect\citeauthoryear{Wang \bgroup \em et al.\egroup
  }{2022}]{wang2022contrastive}
Ran Wang, Xinyu Dai, et~al.
\newblock Contrastive learning-enhanced nearest neighbor mechanism for
  multi-label text classification.
\newblock In {\em Proceedings of the 60th Annual Meeting of the Association for
  Computational Linguistics (Volume 2: Short Papers)}, pages 672--679, 2022.

\bibitem[\protect\citeauthoryear{Xu \bgroup \em et al.\egroup
  }{2019}]{xu2019label}
Ning Xu, Yun-Peng Liu, and Xin Geng.
\newblock Label enhancement for label distribution learning.
\newblock {\em IEEE Transactions on Knowledge and Data Engineering},
  33(4):1632--1643, 2019.

\bibitem[\protect\citeauthoryear{Xu \bgroup \em et al.\egroup }{2021}]{8868206}
N.~Xu, Y.~Liu, and X.~Geng.
\newblock Label enhancement for label distribution learning.
\newblock {\em IEEE Transactions on Knowledge; Data Engineering},
  33(04):1632--1643, apr 2021.

\bibitem[\protect\citeauthoryear{Xu \bgroup \em et al.\egroup
  }{2022}]{xu2022variational}
Ning Xu, Jun Shu, Renyi Zheng, Xin Geng, Deyu Meng, and Min-Ling Zhang.
\newblock Variational label enhancement.
\newblock {\em IEEE Transactions on Pattern Analysis \& Machine Intelligence},
  (01):1--15, 2022.

\bibitem[\protect\citeauthoryear{Yan \bgroup \em et al.\egroup
  }{2016}]{yan2016multi}
Yan Yan, Xu-Cheng Yin, Chun Yang, Bo-Wen Zhang, and Hong-Wei Hao.
\newblock Multi-label ranking with $lstm^2$ for document classification.
\newblock In {\em Chinese Conference on Pattern Recognition}, pages 349--363.
  Springer, 2016.

\bibitem[\protect\citeauthoryear{Yin \bgroup \em et al.\egroup
  }{2006}]{yin20063d}
Lijun Yin, Xiaozhou Wei, Yi~Sun, Jun Wang, and Matthew~J Rosato.
\newblock A 3d facial expression database for facial behavior research.
\newblock In {\em 7th international conference on automatic face and gesture
  recognition (FGR06)}, pages 211--216. IEEE, 2006.

\bibitem[\protect\citeauthoryear{Zhang \bgroup \em et al.\egroup
  }{2015}]{zhang2015crowd}
Zhaoxiang Zhang, Mo~Wang, and Xin Geng.
\newblock Crowd counting in public video surveillance by label distribution
  learning.
\newblock {\em Neurocomputing}, 166:151--163, 2015.

\bibitem[\protect\citeauthoryear{Zhang \bgroup \em et al.\egroup
  }{2022}]{zhang2022use}
Shu Zhang, Ran Xu, Caiming Xiong, and Chetan Ramaiah.
\newblock Use all the labels: A hierarchical multi-label contrastive learning
  framework.
\newblock In {\em Proceedings of the IEEE/CVF Conference on Computer Vision and
  Pattern Recognition}, pages 16660--16669, 2022.

\bibitem[\protect\citeauthoryear{Zhao \bgroup \em et al.\egroup
  }{2022}]{ijcai2022p524}
Xingyu Zhao, Yuexuan An, Ning Xu, and Xin Geng.
\newblock Fusion label enhancement for multi-label learning.
\newblock In Lud~De Raedt, editor, {\em Proceedings of the Thirty-First
  International Joint Conference on Artificial Intelligence, {IJCAI-22}}, pages
  3773--3779. International Joint Conferences on Artificial Intelligence
  Organization, 7 2022.
\newblock Main Track.

\bibitem[\protect\citeauthoryear{Zheng \bgroup \em et al.\egroup
  }{2021}]{zheng2021generalized}
Qinghai Zheng, Jihua Zhu, Haoyu Tang, Xinyuan Liu, Zhongyu Li, and Huimin Lu.
\newblock Generalized label enhancement with sample correlations.
\newblock {\em IEEE Transactions on Knowledge and Data Engineering}, 2021.

\end{thebibliography}

\end{document}